# Framework para Caracterizar Fake News en Términos de Emociones

## Framework for Characterizing Fake News in Terms of Emotions

Luis Rojas Rubio1*
Claudio Meneses Villegas1

1 Universidad Católica del Norte. Departamento de Ingeniería de Sistemas y Computación. Antofagasta, Chile. E-mail: larr014@gmail.com; cmeneses@ucn.cl
* Autor de correspondencia: larr014@gmail.com

**ABSTRACT:**
Las redes sociales se han convertido en uno de los principales canales de información del ser humano debido a la inmediatez e interactividad social que ofrecen, permitiendo en algunos casos publicar lo que cada usuario considere pertinente. Esto ha traído consigo la generación de noticias falsas o Fake News, publicaciones que solo buscan generar incertidumbre, desinformación o sesgar la opinión de los lectores. Se ha evidenciado que el ser humano no es capaz de identificar en su totalidad si un artículo es realmente un hecho o bien una Fake News, debido a esto es que surgen modelos que buscan caracterizar e identificar artículos basados en minería de datos y machine learning. Este artículo propone un framework de tres capas, cuyo principal objetivo es caracterizar las emociones presentes en Fake News y ser una herramienta para trabajos futuros que permitan identificar el estado emocional y estado intencional de quien pública.
**Keywords**: Fake News, Emotions, Framework.

# INTRODUCTION

El uso de redes sociales y la presencia de Fake News han generado un impacto negativo en la sociedad [1], generando incertidumbre y desinformación, anexo a esto se ha evidenciado que el ser humano no es capaz de identificar en su totalidad si un artículo es realmente un hecho o bien un Fake News, debido a esto diversos trabajan a buscado caracterizar las Fake News.

El estado del arte da cuenta de trabajos que buscan caracterizar de diversas formas las Fake News: características textuales [2], tiempos verbales [3], complejidad cognitiva [4], categorización de posturas [5], polaridad emocional [6][7], entre otros.

La contribución de este artículo se basa en la formulación de un framework para la identificación de emociones presentes en publicaciones de redes sociales. Llevado al ámbito de las Fake News, este framework permite caracterizar no solo las emociones presentes, sino también como estas varían respecto a publicaciones noFake.

# FRAMEWORK

El Framework propuesto (Ver Figura 1) busca identificar emociones en un conjunto de publicaciones, para ello se propone la aplicación de tres capas: Emocionalización, Cuantificación y Caracterización. Cada una de estas capas trabaja en función de las palabras utilizadas, el tipo de publicación y el dominio al que pertenecen.

**Fuente**: Elaboración propia.

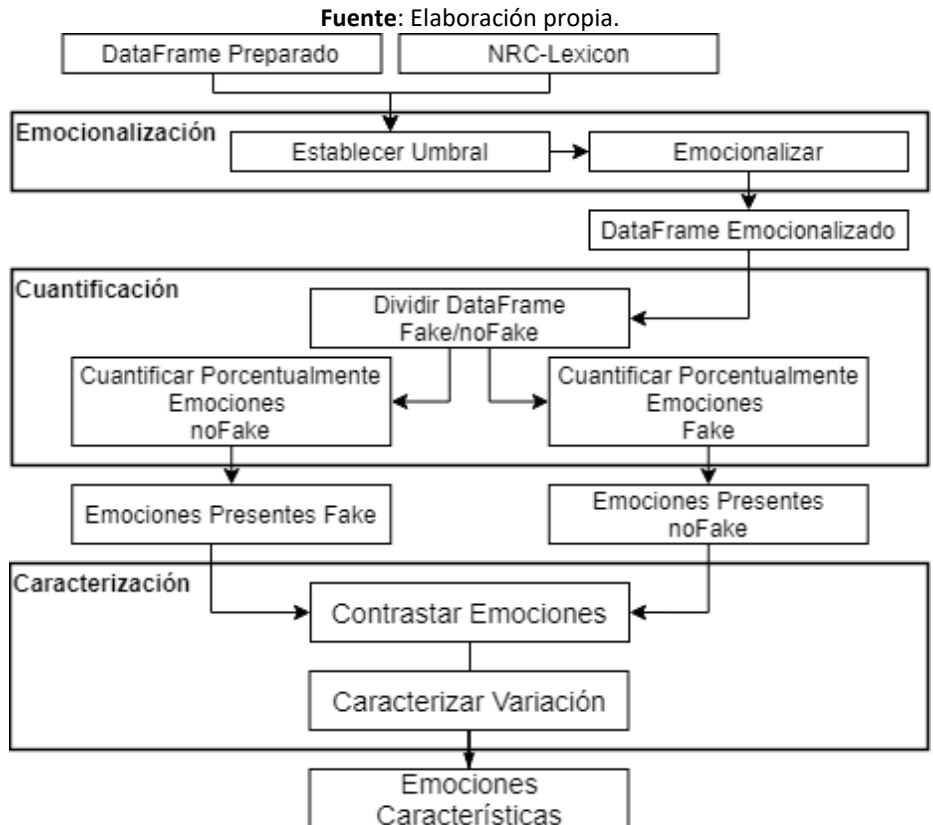

**Figura 1**. Framework

## Emocionalización

La primera capa del framework propuesto requiere de dos conjuntos de datos, NRC-Emotion Lexicon, un dataset presentado en [8] el cual incorpora palabras, emociones y la intensidad con la cual se relacionan. El segundo conjunto de datos debe estar relacionado a las publicaciones que serán analizadas en función de emociones. Se espera una fase previa de preparación de datos para dicho conjunto, exceptuando la aplicación de Lematización y Stemming.

En el 2019 Anoop, Deepak y Lajish presentan una forma de incorporar emociones a Fake News relacionados al área de la salud con el propósito de mejorar sus modelos de clasificación [9]. El método implementado busca comparar cada palabra de una publicación con NRC-Emotion Lexicon. La emocionalización puede ser representada con la ecuación 1, donde $W$ es una palabra de una publicación, $E$ la emoción con la cual podría estar relacionada e *I* la intensidad con la que estas se relacionan, de acuerdo a esto se establece un umbral (*τ*) de modo que, si la intensidad de la emoción respecto a la palabra dada resulta ser mayor o igual a *τ* se procederá a retornar al corpus tanto la palabra *W* como la emoción *E*, en caso contrario solo se mantendrá *W*.

$$f_{(W,E,I)} = \begin{cases} W + " " + E & I \geq \tau \\ W & I < \tau \end{cases} \quad (1)$$

### Cuantificación

La capa de cuantificación inicia con el DataFrame emocionalizado de la capa anterior, el cual es dividido en dos nuevos conjuntos en razón al tipo de publicación; Fake y noFake. Posterior a ello cada DataFrame es sometido a un proceso de cuantificación, dicha cuantificación se puede expresar de acuerdo a la ecuación 2, donde donde $E_{(e,i)}$ representa la frecuencia de una emoción (*e*) en un tipo de publicación (*i*) y $W_{(i)}$ el número de palabras por tipo de publicación (*i*). De ser finalizada esta capa ya se pueden apreciar las emociones presentes por tipo de publicación.

$$P_{(e,i)} = \frac{E_{(e,i)} * 100}{W_{(i)}} \quad (2)$$

### Caracterización

La caracterización es la ultima capa de nuestro framework, la cual se compone de dos procesos: Contraste de emociones y Caracterización de variaciones. El contraste de emociones puede ser representado con la ecuación 3, donde la variación de una emoción ($V_{(e)}$) puede ser calculada como la diferencia de la cuantificación porcentual de la emoción *e* para publicaciones noFake ($P_{(e,'noFake')}$) y la cuantificación porcentual de la misma emoción *e* para publicaciones Fake ($P_{(e,'Fake')}$).

$$V_{(e)} = P_{(e,'noFake')} - P_{(e,'Fake')} \quad (3)$$

Por otra parte, la caracterización de variaciones permite identificar de manera gráfica las variaciones detectadas. Para ello se dispone la ecuación 4, donde:

$$C_{(e,\mu)} = \begin{cases} \downarrow & V_{(e)} > \mu \\ \uparrow & V_{(e)} < -\mu \\ = & \mu \geq V_{(e)} \geq -\mu \end{cases} \bigg| \mu > 0 \quad (4)$$

- Si $V_{(e)} > \mu$, se puede traducir como que una emoción *e* se presenta menos en publicaciones Fake.
- Si $V_{(e)} < -\mu$, se traduce como que una emoción *e* se presenta más en publicaciones Fake.

- Si $\mu \geq V_{(e)} \geq -\mu$, indica que una emoción en se presenta en igual medida en publicaciones Fake como en noFake.

# APLICACIÓN

A modo de validar el uso del framework presentado, se recopilan tres conjuntos de datos: PolitiFact, GossipCop y CoronaFake. Donde el primer conjunto de datos se relaciona al dominio de la política, el segundo a celebridades y el ultimo, al Covid-19. Los tres conjuntos abordan tanto publicaciones Fake como noFake, siendo estas recopiladas de Twitter. Cabe destacar que los tres conjuntos de datos han sido previamente preparados, eliminando URLs, convirtiendo todo a minúsculas, eliminando las StopWords y removiendo los duplicados.

En la tabla [1] se puede apreciar el resultado de la cuantificación ($P_{(e,i)}$), evidenciando cuando frecuente es una emoción respecto a la totalidad de las palabras y tipo de publicación. En negritas se presenta el tipo de publicación de un dominio determinado donde una emoción es más frecuente. Por otra parte, se encuentran subrayados el tipo de publicación del dominio donde la emoción es más frecuente.

**Tabla 1**. Resultados Cuantificación por dominio y tipo de publicación.

| Emoción | Política | | Celebridades | | Covid-19 | |
|---|---|---|---|---|---|---|
| | noFake | Fake | noFake | Fake | noFake | Fake |
| Anger | 0.57% | **<u>1.11%</u>** | 0.72% | **0.90%** | 0.33% | **0.77%** |
| Anticipation | 0.42% | 0.42% | **<u>1.32%</u>** | 0.90% | 0.45% | **0.49%** |
| Disgust | 0.32% | **0.93%** | **0.42%** | 0.38% | **<u>1.05%</u>** | 1.00% |
| Fear | 0.77% | **1.84%** | **1.20%** | 1.02% | 2.56% | **<u>3.05%</u>** |
| Joy | **0.83%** | 0.69% | **<u>3.29%</u>** | 2.22% | 0.28% | **0.37%** |
| Sadness | 0.48% | **1.33%** | **1.01%** | 0.96% | 1.63% | **<u>1.79%</u>** |
| Surprise | 0.20% | **0.32%** | **<u>0.5%</u>** | 0.31% | 0.34% | **0.47%** |

**Fuente**: Elaboración propia.

En la tabla [2] se demuestra la aplicación de la capa de caracterización, evidenciando el cambio de las emociones presentes entre noticias Fake y noFake de cada dominio.

**Tabla 2**. Resultados Caracterización por dominio.

| Emoción | Política | Celebridades | Covid-19 |
|---|---|---|---|
| Anger | ↑ | ↑ | ↑ |
| Anticipation | = | ↓ | ↓ |
| Disgust | ↑ | ↓ | ↓ |
| Fear | ↑ | ↓ | ↑ |
| Joy | ↓ | ↓ | ↑ |
| Sadness | ↑ | ↓ | ↑ |
| Surprise | ↑ | ↓ | ↑ |

**Fuente**: Elaboración propia.

# CONCLUSIÓN Y TRABAJO FUTURO

En este artículo se ha propuesto un framework para la identificación de emociones en Fake News de un dominio determinado y se ha presentado su aplicación en tres dominios distintos. Este framework podría traer beneficios tales como, mejoras en las tasas de clasificación de Fake News, detección temprana de Fake News, identificar el estado emocional y/o estado intencional de quien pública.

Como trabajo futuro se espera incorporar al framework herramientas que permitan identificar no solo las emociones, sino también las intensiones de quien publica, buscando establecer relaciones entre las emociones e intenciones.